%% file: main.tex
\documentclass[10pt,twocolumn,letterpaper]{article}

\usepackage{cvpr}              

\input{preamble}

\definecolor{cvprblue}{rgb}{0.21,0.49,0.74}
\usepackage[pagebackref,breaklinks,colorlinks,allcolors=cvprblue]{hyperref}
\usepackage{float}
\usepackage{duckuments}
\usepackage{multirow}

\def\paperID{5860} 
\def\confName{CVPR}
\def\confYear{2025}

\newcommand{\method}{Toon3D}
\title{\method: Seeing Cartoons from New Perspectives}

\author{%
	Ethan Weber\textsuperscript{*2}\qquad%
	Riley Peterlinz\textsuperscript{*2}\qquad%
	Rohan Mathur\textsuperscript{2}\qquad%
	\\[0.15em]%
	Frederik Warburg\textsuperscript{1}\qquad%
	  Alexei A. Efros\textsuperscript{2}\qquad%
	Angjoo Kanazawa\textsuperscript{2}%
	\\[0.5em]%
	\textsuperscript{*}Equal contribution\quad%
	\textsuperscript{1}Teton.ai\quad%
	\textsuperscript{2}UC Berkeley%
}
\DeclareMathOperator*{\argmin}{arg\,min}

\begin{document}

\input{macros}
\input{paper}

{
    \small
    \bibliographystyle{ieeenat_fullname}
    \bibliography{main}
}
\clearpage
\input{appendix}

\end{document}

%% file: preamble.tex
\newcommand{\new}[1]{{\color{black}#1}}


%% file: macros.tex
\newcommand{\ethan}[1]{\textcolor{orange}{ETHAN: #1}}
\newcommand{\riley}[1]{\textcolor{blue}{RILEY: #1}}

\newcommand{\mytilde}{\~{}}

%% file: paper.tex
\twocolumn[{
\renewcommand\twocolumn[1][]{#1}
\maketitle
\centering
\vspace{-2em}
\includegraphics[width=\linewidth]{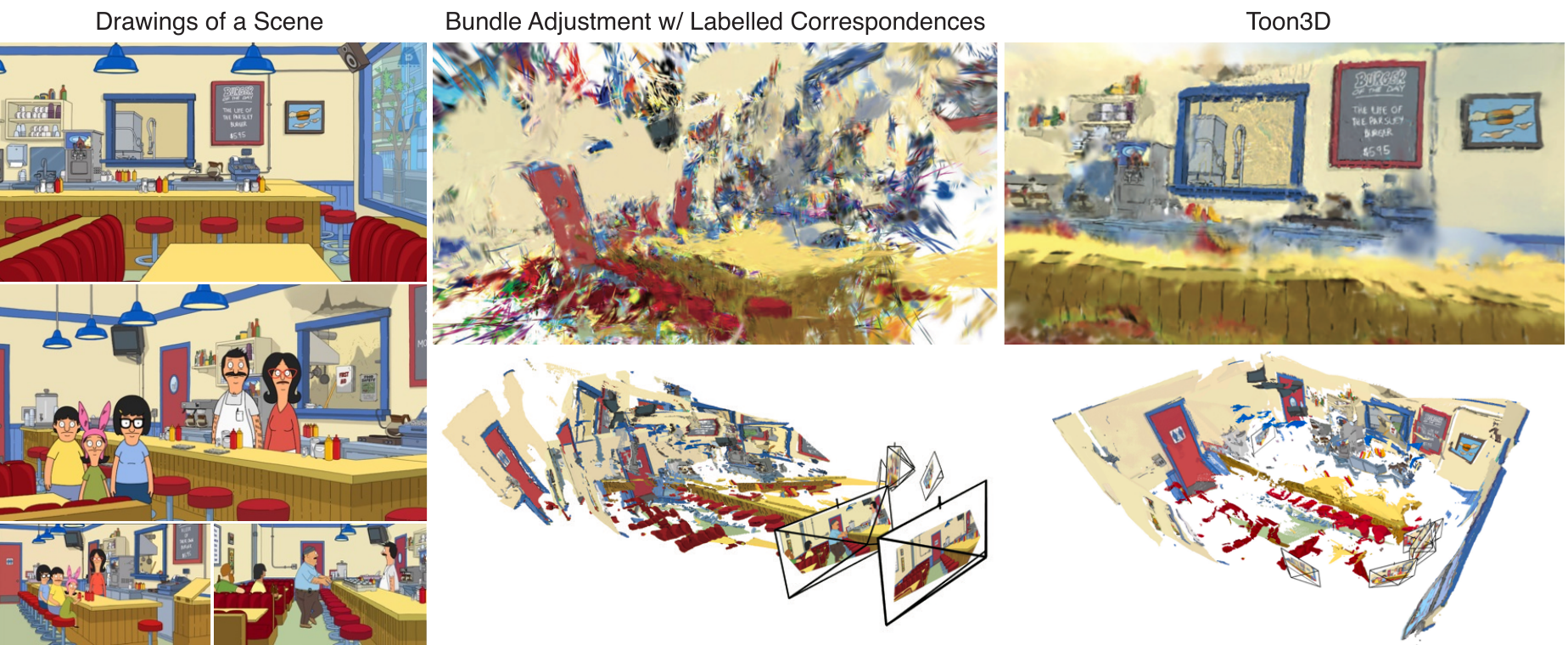}
\vspace{-15px}
\captionof{figure}{\textbf{Reconstructing a 3D scene from 3D inconsistent images.}
Cartoons and animations often depict scenes that are not geometrically consistent by design (left), making it challenging for classical Structure-from-Motion (SfM) techniques to reconstruct these scenes as they assume 3D consistency (middle). However, humans can easily perceive the underlying 3D scene from these images. We introduce Toon3D, which addresses these challenges by deforming images during reconstruction to account for geometric inconsistencies and leveraging monocular depth priors. The middle column illustrates how Bundle Adjustment fails, even with manually labeled correspondences, resulting in scattered Gaussian splats (top) and misaligned camera reconstructions visualized by backprojected monodepths (bottom). The right column shows our Toon3D results, with more coherent Gaussian splats (top) and well-structured point clouds and camera views (bottom), demonstrating significantly improved 3D consistency. Our project page is \url{https://toon3d.studio/}.
}
\label{fig:teaser}
\vspace{1em}
}]

\begin{abstract}
We recover the underlying 3D structure from images of cartoons and anime depicting the same scene.
This is an interesting problem domain because images in creative media are often depicted without explicit geometric consistency for storytelling and creative expression—they are only 3D in a qualitative sense. While humans can easily perceive the underlying 3D scene from these images, existing Structure-from-Motion (SfM) methods that assume 3D consistency fail catastrophically.
We present Toon3D for reconstructing geometrically inconsistent images.
Our key insight is to deform the input images while recovering camera poses and scene geometry, effectively explaining away geometrical inconsistencies to achieve consistency. This process is guided by the structure inferred from monocular depth predictions.
We curate a dataset with multi-view imagery from cartoons and anime that we annotate with reliable sparse correspondences using our user-friendly annotation tool.
Our recovered point clouds can be plugged into novel-view synthesis methods to experience cartoons from viewpoints never drawn before. We evaluate against classical and recent learning-based SfM methods, where Toon3D is able to obtain more reliable camera poses and scene geometry.
\end{abstract}

\vspace{-20px}
\section{Introduction}
\label{sec:introduction}

Humans typically have little trouble inferring the relative camera poses and 3D structure from hand-drawn cartoons. However, current structure-from-motion (SfM) pipelines fail to reconstruct these scenes because (1) the images are not geometrically consistent, (2) the images do not obey physically plausible camera models, (3) the scenes are typically only drawn from a sparse set of views, and additionally, (4) many outlier correspondences from automatic methods. In this work, we overcome these challenges by proposing a piecewise-rigid deformable optimization framework that recovers camera poses and 3D scene from geometrically inconsistent images (see Fig.~\ref{fig:teaser}). 

\begin{figure*}[t]
\centering
\includegraphics[width=\linewidth]{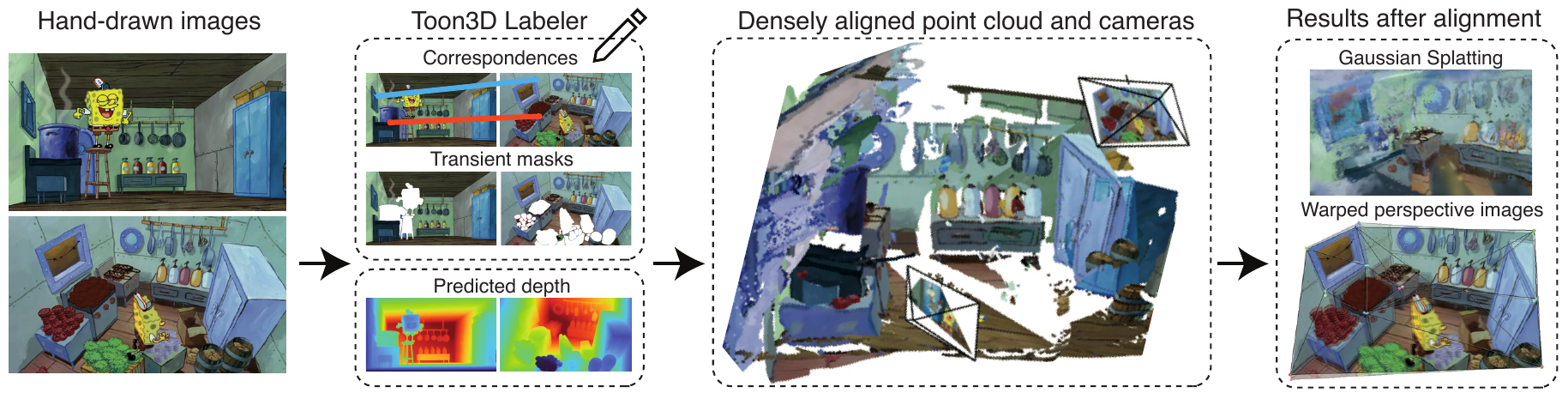}
\vspace{-15pt}
\caption{\textbf{Toon3D overview.} Our framework consists of labeling images with our interactive Toon3D Labeler tool, recovering camera poses and aligning a dense point cloud, and visualizing the dense reconstruction with Gaussians to create an immersive visual experience.}%
\label{fig:method}
\vspace{-1.5em}
\end{figure*}

Our pipeline consists of a joint optimization to recover cameras and aligned geometry. It takes a set of correspondences as input, which we backproject into 3D using the depth from a monodepth network~\cite{ke2023repurposing_marigold,wang2024moge}. We align these sparse correspondences in 3D to estimate the camera intrinsic and extrinsic parameters. Simultaneously, we also deform the image and the associated depth such that images satisfy 3D consistency.
We regularize our warps with 2D and 3D rigidity losses to prevent degenerate solutions. 

We also propose the Toon3D Dataset and the Toon3D Labeler which is a user-friendly annotation tool, where a user can label point correspondences between images while segmenting transient objects. The Toon3D Labeler is a hosted website with no installation, so anyone can get up and running with it easily. We intentionally highlight Toon3D Labeler as a contribution of our paper because artists work with cartoon drawings regularly, and this tool fits nicely into a human-in-the-loop framework for recovering 3D from these drawings. Our recovered 3D model may help artists draw novel viewpoints. We use our labeler to label 12 scenes from popular cartoons and anime, such as Sponge Bob (Fig.~\ref{fig:teaser}) and Spirited Away, and we release these as the Toon3D Dataset.

To the best of our knowledge, we are the first to present a pipeline for reconstructing cartoon or hand-drawn scenes. Our pipeline yields reliable camera poses, whereas COLMAP~\cite{schoenberger2016sfm} and DUSt3R~\cite{wang2024dust3r} fails to recover camera poses and 3D scene geometry (even with human-annotated correspondences) due to 3D inconsistencies in the input images. In contrast, our 2D image warpings of the original images enable us to reconstruct the full 3D geometry, while also visualize geometrical inconsistencies in the drawings.

We evaluate our pipeline on 12 popular scenes (10 cartoon TV shows, 1 movie) to highlight the effectiveness of our pipeline in obtaining good camera poses and reconstructions. We show reconstructions of our recovered 3D point clouds and create an immersive visualization by rendering a 3D Gaussian Splatting~\cite{kerbl3Dgaussians} representation that are initialized from our aligned point cloud. We evaluate our proposed alignment objectives and losses qualitatively and quantitatively. We demonstrate that our warps can highlight geometric inconsistencies in hand-drawn images. We further validate the quality of Toon3D to estimate camera poses, when the scenes are in fact geometrically consistent. We show that we can obtain the 3D geometry of Airbnb rooms with sparse views. Finally, we show that Toon3D is also useful for reconstructing the 3D geometry from paintings depicting the same landmark from different views.

Humans routinely make successful 3D scene inferences from imagery (e.g. cartoons) which is 3D-inconsistent and/or not following perspective projection~\cite{hertzmann2024toward}.   
Toon3D is a step toward achieving this type of  qualitative 3D understanding of cartoons. 
We validate our pipeline and will release all data, code, and tools to easily process any cartoon. We hope our contribution serves as a useful framework to build tools that, like humans, can reconstruct and understand qualitative 3D.

\section{Related work}
\label{sec:related_work}

\textbf{Multi-view geometry estimation.} Structure-from-Motion (SfM)~\cite{hartley2003multiple,snavely2006photo} takes in images, detects and matches correspondences, and solves for camera parameters. COLMAP~\cite{schoenberger2016sfm} is a popular SfM pipeline, but it fails for wide baseline images (few correspondences), images with a lot of moving objects, or geometric inconsistencies typically present in cartoons. Improvements in keypoint detection~\cite{detone2018superpoint, dusmanu2019d2}, matching~\cite{sun2021loftr, sarlin2020superglue} and optimizations~\cite{tirado2023dac} have been proposed to better handle wide baselines~\cite{vallone2022danish} and be robust to transient objects~\cite{bescos2018dynaslam}. However, all these methods make a fundamental assumption that the input images are geometrically consistent. In contrast, we propose a method that accounts for such inconsistencies by explaining away the inconsistencies when possible via image deformation.

\vspace{.5em}\noindent \textbf{Reconstructing image collections.} 
Facade~\cite{debevec1996modeling}, a seminal early work in image-based modeling and rendering, used a set of photographs of an architectural scene to recover a textured 3D model using structure-from-motion with human-specified volumetric constraints. Phototourism~\cite{snavely2006photo} and Building Rome in a Day~\cite{AarwalBuildingRome} pioneered the use of large online photo collections for 3D reconstruction.
Object-centric methods like CMR~\cite{kanazawa2018learning,kanazawa2016learning} recover 3D models of animals through a learned deformation model.
For non-rigid dynamic scenes, there exist methods which explain small variations in a video via a 3D model with a time-conditioned warp fied to be as rigid as possible~\cite{park2021nerfies, tretschk2021non, pumarola2021d}. With methods that require deformation, techniques such as As-Rigid-As-Possible (ARAP)~\cite{sorkine2007rigid} are useful. These problems are relevant in a sense that they need to reconstruct scenes with transient variations in each image. We propose a relevant but novel and under-explored problem setting where the input images are meant to depict the same 3D scene, through geometrically inconsistent multi-view imagery.

\vspace{.5em}\noindent\textbf{Paintings to 3D.} Most attempts at recovering 3D from drawings and paintings have focused on the single view setting, with missing 3D information provided either manually by the user or via learning.  Important early user-assisted approaches for generating 3D scenes from a single painting include Tour into the Picture~\cite{horry1997tour}, which assumed single-point perspective, and the more general Single View Metrology~\cite{criminisi2000single}.
Automatic Photo Popup~\cite{hoiem2005automatic} replaced the manual parts of the reconstruction process with early machine learning techniques, and was able to generalize to paintings. Aubry \etal~\cite{Aubry13} is a rare attempt to connect different paintings of the same scene by using a 3D model.  There has also been a few attempts to recover a 3D model from a set of sketches of the same object~\cite{delanoy20183d,guillard2021sketch2mesh}. Our approach similarly explores reconstructing creative expressions (\ie drawings) but from multiple drawings of the same scene as seen in settings like cartoons instead of a single image. 

\vspace{.5em}\noindent\textbf{Computer vision in TV and Film.}
Previous works have explored reconstructing TV shows and films.  Pavlakos \etal~\cite{pavlakos2022sitcoms3D} recover camera shot locations, 3D human poses, and gaze understanding, enabling applications such as post-production re-rendering with novel camera paths. 
MovieNet~\cite{huang2020movienet} proposes a large dataset of popular films annotated with bounding boxes, actions, and cinematic style for a holistic understanding of movies. 
Zhu \etal~\cite{zhu2015aligning} align movies and books to obtain fine-grained descriptions of appearances of objects and characters, as well as high-level semantic understanding into how characters think and reason. Additionally, some works have looked into character reconstruction for cartoon characters~\cite{chen2023panic3d,jain2012three,10.1145/3592788} but none have looked at recovering camera poses and reconstructing full 3D environments. Our work is most similar to~\cite{pavlakos2022sitcoms3D}, but we tackle geometrical inconsistencies in cartoons and animation instead of video sequences in sitcoms.

\begin{figure*}[t]
\centering
\includegraphics[width=\linewidth]{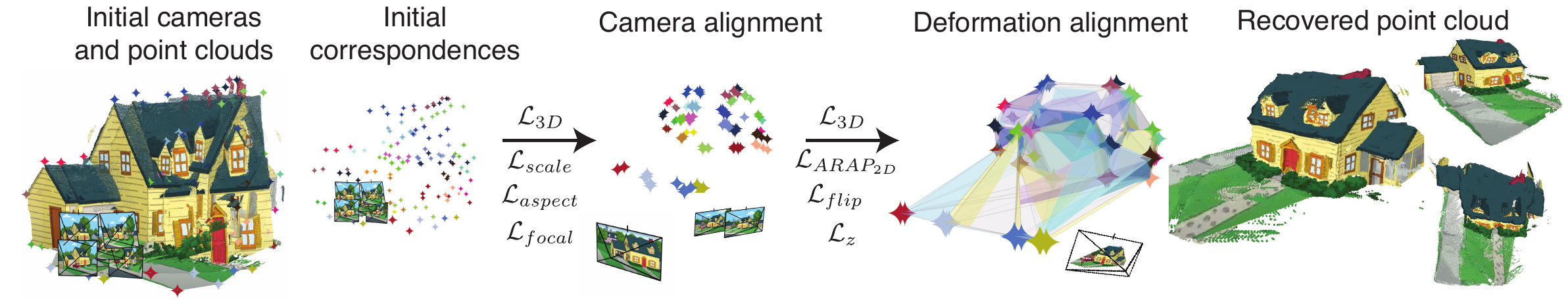}
\vspace{-15pt}
\caption{\textbf{Toon3D alignment.} The camera alignment objective aligns the point clouds while optimizing for camera intrinsics and extrinsics. Deformation alignment deforms the images to obey a perspective camera model. In practice, our method uses all the losses described here to obtain an aligned point cloud and posed images.}
\vspace{-1em}
\label{fig:alignment}
\end{figure*}

\vspace{-5px}
\section{Toon3D Dataset and Labeler}
\label{sec:dataset}
To study this unique problem, we introduce the Toon3D Dataset, which consists of 12 cartoon scenes (10 TV shows, 1 movie) each with 5-12 images depicting the same environment.
An innate challenge in cartoons is that correspondences are difficult to obtain automatically. We tried several \new{SOTA} keypoint detectors~\cite{sun2021loftr, sarlin2020superglue, detone2018superpoint, amir2021deep}, 
but they often fail due to extreme viewpoint changes, the presence of transient objects such as characters, and the images’ stylistic, low-texture expression. Since our focus is on reconstructing the underlying static 3D scene, we leave the automatic removal of foreground objects and estimation of 2D correspondences to future work. Instead, we develop the Toon3D Labeler, a human-in-the-loop tool for segmenting transient objects and annotating sparse 2D correspondences. The Toon3D Labeler is hosted online with no installation required, making it easily accessible. See the appendix or project page for a visualization of this tool. Next, we discuss how we use the Toon3D Labeler to curate our dataset.

\noindent\textbf{Preprocessing.} We start with a set of $N$ images $\{I_i\}$ depicting the same scene in a cartoon. Each scene typically has $N \le 10$ images with wide baselines. We preprocess these images by running a monocular depth network to obtain predicted depths $\{D_i\}$.
\new{We normalize the depth maps by dividing by the maximum depth of a labeled correspondence across all depth maps.}
We experiment with a variety of depth map predictors ~\cite{ke2023repurposing_marigold,wang2024moge,yang2024depth}, while all quantitative evaluations are done with Marigold~\cite{ke2023repurposing_marigold}. \new{We also} run Segment Anything (SAM)~\cite{kirillov2023segment} to get a set of masks per image.

\noindent\textbf{Labeling.} We label these images using the Toon3D Labeler on the web interface.
To annotate correspondences, the user clicks on corresponding points across all images. When the point is not visible in an image, it is labeled as invisible. Our interface allows users to visualize the depth map, helping them avoid placing correspondences on depth discontinuities. Each annotated image has on average 18 sparse correspondences (\new{see more details in appendix}).
To select SAM mask, the user simply hovers over a region, the mask will be highlighted, and it can be toggled on and off to discard those transient pixels.  \new{After labeling, we have} pixel correspondences $\mathcal{X} = \{x_{i,c}\}$ where $i$ is the image index and $c$ is the correspondence index. We also have a valid correspondences mask $m_{i,c}=\{0,1\}$. When $m_{i,c}=0$, the correspondence is not visible in that image. We denote the predicted depth of the correspondences with $d_{i,c} = D_{i}(x_{i,c})$.

\section{Toon3D Method}
\label{sec:method}

We present Toon3D, a method to reconstruct scenes that are only 3D consistent in a qualitative sense, as opposed to existing SfM methods that requires a geometrical consistent scene. Toon3D takes as input multiple images of the same scene with point correspondences, mask annotations, and estimated monocular depth, and outputs camera poses for each image, a 3D point cloud, and a warping of the original images, such that they obey a perspective camera model. The output point cloud can be converted into a Gaussian Splatting~\cite{kerbl3Dgaussians} representation to create a more immersive novel-view experience. We optimize for cameras and geometry by aligning backprojected point correspondences and allowing the images to deform while still obeying a perspective camera model and multiview geometry. We will now explain our approach in more detail.

\vspace{-10px}
\subsection{Camera Alignment}
\vspace{-5px}
\label{sec:sparse_alignment}
The first objective of our pipeline is to obtain camera poses. Since the images are not geometrically consistent, the standard bundle adjustment process that enforces a single 3D point for every corresponding points does not lead to correct camera poses. Instead, we make use of the monocular depth priors in each image and solve for camera poses that aligns the backprojected correspondences in 3D.

\begin{figure*}[t]
\centering
\includegraphics[width=\linewidth]{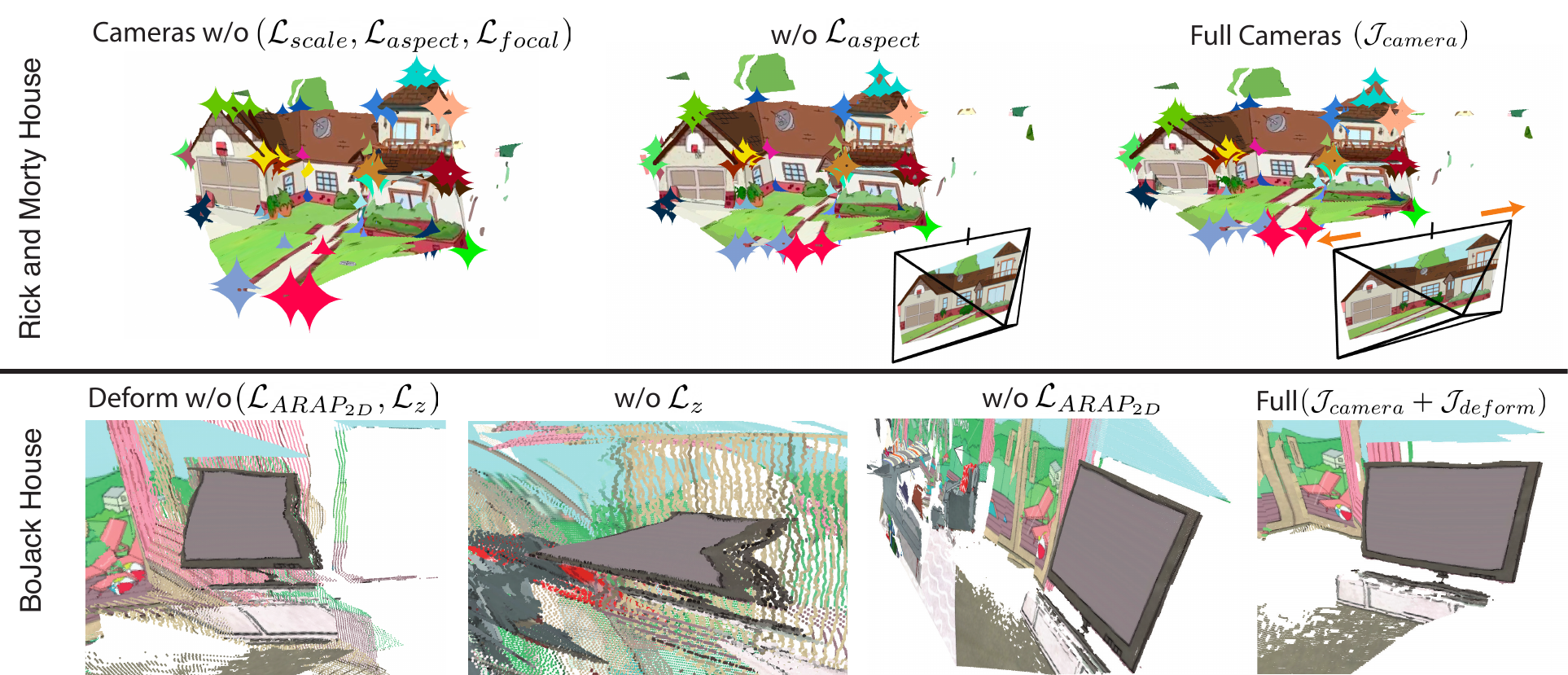}
\vspace{-20pt}
\caption{\textbf{3D alignment ablations.} Row 1 (Rick and Morty House) shows regularization's impact on scene shaping. Optimized shift and scale parameters can adjust point clouds to better align at correspondences. This is evident as the starred points converge. The aspect regularization keeps the optimized image close to its original aspect ratio. Row 2 (BoJack Horseman House) explores the effects of different warp regularizers ($\mathcal{L}_{ARAP_{2D}}$ and $\mathcal{L}_{z}$) on scene warping. Without any regularization, warping distorts scene geometry. ARAP alone results in poor 3D warps due to inaccurate depth. $z$ regularization alone limits scene movement, maintaining rigid structures close to the original depth map. Using both strikes a good balance between correctly positioning geometry and preserving structural integrity.}
\label{fig:ablations}
\vspace{-1em}
\end{figure*}

Specifically, we first backproject our sparse correspondences into 3D with
\begin{align}
    p(x_{i,c}) = R_i \cdot K_i^{-1} \cdot \left( s_i \cdot d_{i,c} + h_i \right),
\end{align}
where the depth $d$ of each point is estimated with a monocular depth network and we solve for camera rotations $R$, translations $t$, focal lengths $f_x, f_y$, depth scale $s$, and shift $h$ that minimizes the 3D correspondence loss

\begin{align}
    \mathcal{L}_{3D} = \frac{1}{|\mathcal{X}|} \sum_{i=1}^{N} \sum_{j < i}^{N} \sum_{c=1}^{M} m_{i,c} \cdot || p(x_{i,c}) - p(x_{j,c}) ||_{2}^{2},
\end{align} which pulls the backprojected correspondences together in 3D. We found minimizing 3D distance rather than 2D reprojection error empirically, which we ablate in  experiments.

Estimating these camera poses from just few sparse correspondences is a very under-constrained problem even with a strong depth prior. Therefore, we found that adding the following regularizes were necessary to reliably estimate of camera poses across all scenes.

\vspace{-10pt}
\footnotesize
\begin{align}
\mathcal{L}_{scale}&=|| 1 - \frac{1}{N} \sum_{i=1}^{N} s_{i} ||^2 \\
\mathcal{L}_{aspect}&=\sum_{i=1}^{N} || \frac{f_{i,x}}{f_{i,y}} - \frac{h_i}{w_i}||^2 \\
\mathcal{L}_{focal}&=\sum_{i=1}^{N} f_{i,x} + f_{i,y},
\end{align}
\normalsize
where $\mathcal{L}_{scale}$ encourages a scale close to 1 such that the scene does not shrink, $\mathcal{L}_{aspect}$ balances $f_{i,x}$ and $f_{i,y}$ to maintain aspect ratio of the camera with the original image's height $h_i$ and width $w_i$, and $\mathcal{L}_{focal}$ penalizes large focal length to prefer wide-angle cameras over far away and zoomed in shots. We also have losses that penalize scales $s_i$ and shifts $h_i$ if they become negative with $\mathcal{L}_{neg}(x) = ||\frac{1}{N} \sum_{i=1}^{N} \max(0, -x_i) ||^2$. 

Thus, our final camera alignment objective is as follows

\vspace{-10pt}
\begin{equation}
\begin{split}
\mathcal{J}_{\text{camera}} = & \ \mathcal{L}_{3D} + \lambda_{scale} \mathcal{L}_{scale} + \lambda_{aspect} \mathcal{L}_{aspect} \\
+ & \lambda_{focal} \mathcal{L}_{focal} + \lambda_{neg} (\mathcal{L}_{neg}(s) + \mathcal{L}_{neg}(h)),
\end{split}
\end{equation}
which gives us an coarse estimate of the 3D structure and the camera poses.

\subsection{Deformation Alignment}
\vspace{-5px}
Although the previous losses yield coarse estimates of the scene, they do not result in a coherent point cloud due to the geometric inconsistencies in cartoon images.
To address this, we propose jointly deforming each image and its corresponding depth map to achieve geometric consistency. Our method introduces a set of dense alignment objectives, which, when optimized, refine the camera poses and produce a densely aligned, warped 3D point cloud along with images that are geometrically consistent in 3D and adhere to a perspective camera model.

To do this, we use the same optimization objectives from Sec.~\ref{sec:sparse_alignment} but now with more freedom as we also allow the input image to be warped to further minimize $\mathcal{L}_{3D}$. However, naively warping every pixel location with full degrees of freedom, without any constraints, results in degenerate solutions. To address this, we warp the image using a coarse 3D mesh that approximates the scene and apply a regularizer to ensure the deformation remains piece-wise rigid.

Specifically, we first transform each training image and predicted depth into a 3D mesh with vertices $V \in \mathbb{R}^{M\times3}$ and faces $F \in \mathbb{R}^{K\times3}$, where $V_{i,xy}$ is the initial 2D point for image $i$ and $V_{i,z}$ is the initial depth. We use the  labeled correspondences $x_{i,c}$ as the vertices of this mesh. We use Delaunay triangulation to create the mesh topology. See Fig.~\ref{fig:alignment} for illustrations of this 3D mesh that represents the scene for each image. 

We optimize the $V$ of each image with various 2D and 3D regularizers to constrain the warps to be as-rigid-as-possible to prevent degenerate solutions. First, we regularize such that the optimized vertices  are encouraged to follow a rigid transform in the 2D image plane via

\vspace{-10px}
\begingroup
\footnotesize
\begin{align}
&\mathcal{L}_{ARAP_{2D}} = \frac{1}{N \times |\mathcal{F}|} \sum_{i=1}^{N}
\sum_{f \in F_i} ||\pi(V_{i}^{'}[f]) - A_{i \rightarrow j}\pi(V_{i}[f])||^2,
\end{align}
\endgroup
where $\pi$ denotes the 2D projection with the current camera parameters, $V_{i}[f] \in \mathbb{R}^{2\times3}$ are vertices indexed at face $f$, $V'$ are the optimized 2D projected vertices, and $A_{a \rightarrow b}$ is the best fit 2D rigid transform in the image plane that transforms vertices $V_{i}[f] $ to the new vertices $V_{i}^{'}[f]$.

Additionally, we use these two losses

\begin{figure*}[t]
\centering
\includegraphics[width=\linewidth]{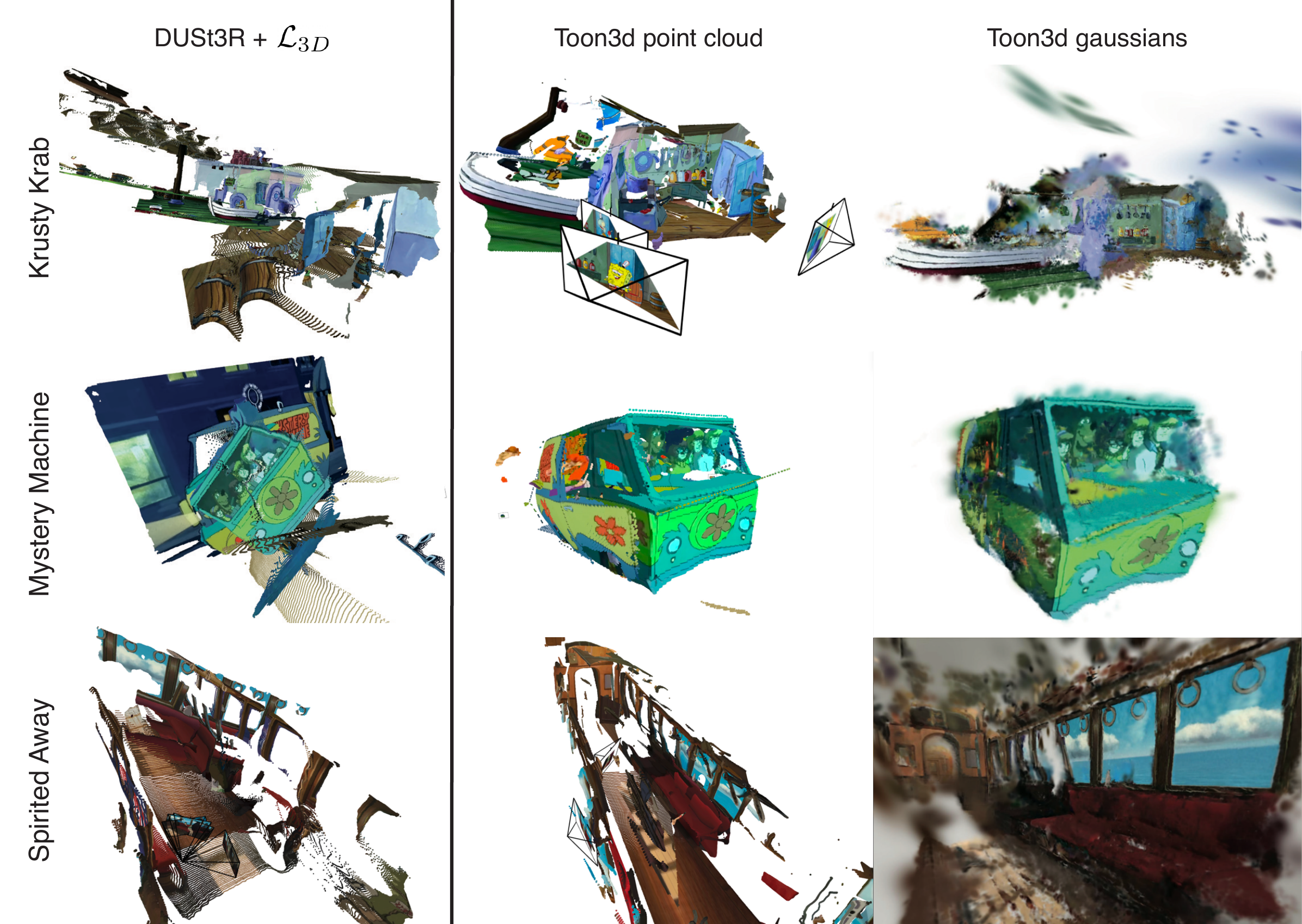}
\vspace{-15pt}
\caption{\textbf{3D reconstructions of cartoons.}
Off-the-shelf methods like COLMAP fail completely. State-of-the-art learning based method \new{DUSt3R~\cite{wang2024dust3r} also fails catastrophically on many scenes even with labeled correspondences (left). } Our method (middle), recovers reliable camera, and plausible pointcloud, which can be visualized with Gaussians for a more immersive experience. 
For the SpongeBob scene (top), we label point correspondences between walls to reconstruct two rooms together. Notably, our method works with different depth predictors. From top to bottom, we show results with MoGe~\cite{wang2024moge}, Depth Anything V2~\cite{yang2024depth}, and Marigold~\cite{ke2023repurposing_marigold}.}
\label{fig:our_results}
\vspace{-1.5em}
\end{figure*}

\vspace{-10px}
\begingroup
\footnotesize
\begin{align}
&\mathcal{L}_{flip} = \frac{1}{N \times |\mathcal{F}|} \sum_{i=1}^{N}
\sum_{f \in F_i} ||\min(0, t_{area} - \det(V_{i}[f]))||^2,
\end{align}
\endgroup
\begingroup
\footnotesize
\begin{align}
&\mathcal{L}_{z} = \frac{1}{N \times |\mathcal{X}|} \sum_{i=1}^{N} \sum_{c=1}^M m_{i,c} \cdot ||d'_{i,c} - d_{i,c}||,
\end{align}
\endgroup
 where $\mathcal{L}_{flip}$ penalizes if the triangle face gets too small or flips, and  $\mathcal{L}_{z}$ encourages the warped depth to be close to the original predicted depth. $t_{area}$ is the minimum area a face can be, and $\det$ gives the signed face area. We set $t_{area}$ to 10\% of the original face area.

Finally, we use barycentric interpolation to densely warp the RGB and depth maps according to our deformed vertices $V'$. We warp the RGB image with barycentric interpolation according to the original vertices $V$ and the deformed mesh $V'$. Similarly, we compute a depth offset and apply it to the original depth images $d_i$ to obtain $d_i'$. 

Our deformation alignment objective becomes an extension of our camera alignment, where besides optimizing for poses, focal lengths, rotation, scale, and shift, we also optimize the mesh topology. Our final objective is

\vspace{-15px}
\begin{align}
\argmin_{R, t, f, s, h, V} \mathcal{J}_{\text{align}} = & \mathcal{J}_{\text{camera}} + \mathcal{J}_{\text{deform}} \\
\mathcal{J}_{\text{deform}} =  \lambda_{ARAP_{2D}} \mathcal{L}_{ARAP_{2D}} +& \lambda_{flip} \mathcal{L}_{flip} + \lambda_{z} \mathcal{L}_{z}
\end{align}

Optimizing this objective results in an accurate 3D poses as well as an image and depth map that obey the perspective camera model as well as global 3D geometry consistency.

\subsection{Gaussian visualization}
\vspace{-5px}

At this point, we have aligned depth maps which are backprojected into a combined 3D point cloud. We could visualize the point cloud as-is, but we find that Gaussian Splatting can create a more immersive experience. Gaussian Splatting~\cite{kerbl3Dgaussians} is typically initialized by a sparse point cloud from COLMAP, but instead, we initialize it with our dense point cloud. We add a few sparse-view regularizers including the ranking loss from \cite{wang2023sparsenerf} (to reconstruct scenes to be consistent with the predicted depth) and a total variation~\cite{zhou2017unsupervised} loss in novel views interpolated between pairs of training views. The transient regions are not ignored in the objective.

\vspace{-8px}
\section{Experiments}
\label{sec:experiments}
\vspace{-5px}
First we show results on cartoon scenes, and then we evaluate our design choices and compare our method with DUSt3R~\cite{wang2024dust3r}, a state-of-the-art learning based 3D reconstruction method. We further test the correctness of our approach on a similar setup but with geometrically consistent photos from an AirBnB listing. We also evaluate our approach on paintings \new{and finally, visualize which parts of the images need to warp to become consistent with each other.}

\begin{figure*}[t]
\centering
\includegraphics[width=\linewidth]{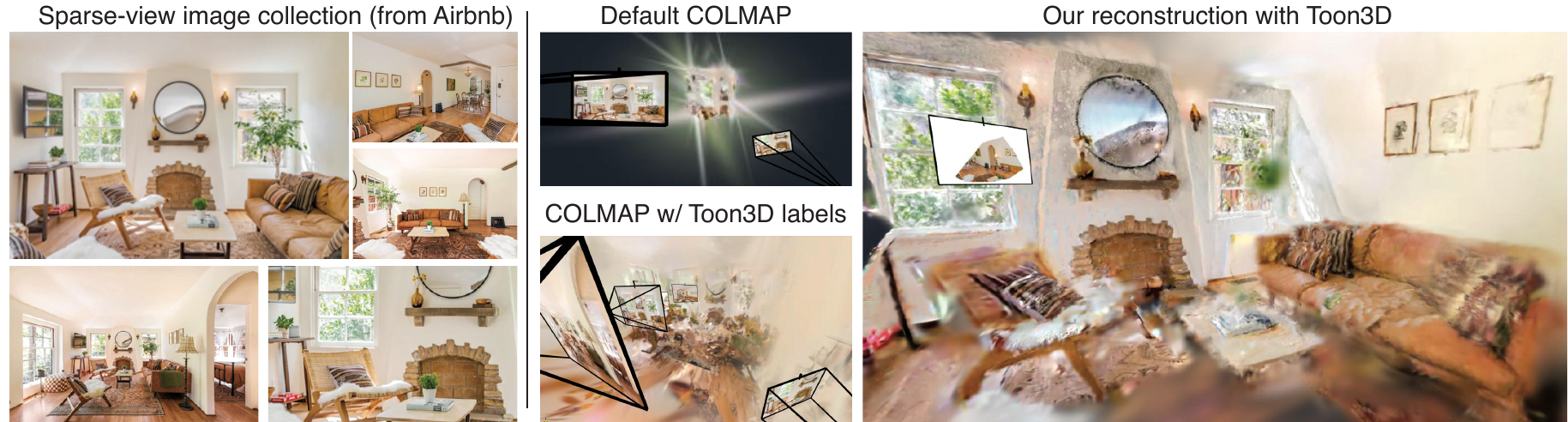}
\vspace{-15px}
\caption{\textbf{Sparse-view Reconstruction.} Our pipeline can reconstruct sparse-view image collections that are geometrically consistent as well (left). COLMAP by default only registers 2 out of 5 images and fails to recover structure (middle top). Using Toon3D Labeler correspondences, we get COLMAP to work (middle bottom) but it is initialized with a very sparse point cloud and cannot recover dense details properly. Using Toon3D, we can fully reconstruct the room.}
\label{fig:sparse-view-reconstruction}
\vspace{-2em}
\end{figure*}

\subsection{Cartoon reconstruction}
\vspace{-5px}
In Fig.~\ref{fig:our_results} we show the results from our pipeline on multiple popular cartoon scenes. \new{On the left, we show results using DUSt3R~\cite{wang2024dust3r}, which often fails catastrophically even with our labeled correspondences. The \new{center} column shows our point cloud reconstruction. The right column shows rendered novel views after the Gaussian visualization. We also show a traditional bundle adjustment (BA) baseline in Fig.~\ref{fig:teaser} that optimizes a single 3D point for each labeled correspondence, which recovers inaccurate poses. For clarity, we visualize the dense result by backprojecting monocular depths. Approaches that don't account for geometrical inconsistencies result in poor camera poses. } \new{Please see our overview video for better visualization.} From start to completion, our method takes on the order of minutes. Finding a few images of a cartoon scene and labeling points is quick due to the web-based viewer, and running our camera alignment and warping takes approximately 1 minute on an NVIDIA RTX A5000. Running Gaussian Splatting in Nerfstudio~\cite{tancik2023nerfstudio} with our additional losses takes $\sim$3 minutes.

\vspace{-10px}
\paragraph{Qualitative ablations.} For our default method, we have all parameters free (including scale and shift) with all regularization losses turned on. We show the qualitative trade-offs for our various losses in Fig.~\ref{fig:ablations}. We find our losses help align structure while maintaining an accurate aspect ratio, preventing degenerate warps, and favoring cameras inside walls rather than far away and zoomed in (see caption).

\vspace{-5px}
\input{evaluation_table}

\begin{figure}[H]
\centering
\includegraphics[width=\linewidth]{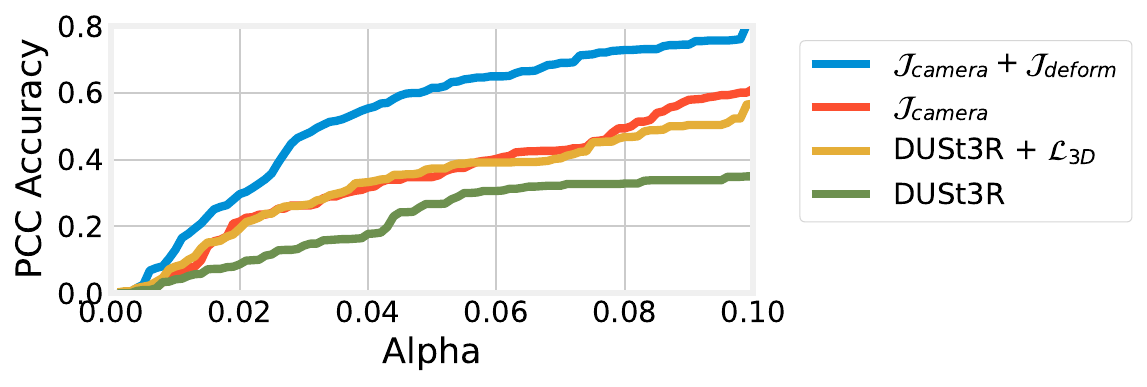}
\vspace{-20px}
\caption{\textbf{Baselines evaluation.} \new{We compare our full method against various baselines. We compare with DUSt3R and improve it with our labeles and $\mathcal{L}_{3D}$. We (blue) obtains best metrics for percent correct correspondences at image size \% thresholds $\alpha$.}}
\label{fig:quantitative-baselines}
\vspace{-1.5em}
\end{figure}

\vspace{-5px}
\paragraph{Quantitative evaluation}
Our task is most naturally evaluated qualitatively, but to be thorough, we design a metric to evaluate 3D consistency, with results reported in Tab.~\ref{tab:ablations}. We randomly remove $5$ labeled correspondence points from each image of our 12 Toon3D scenes and report the average percentage of correct correspondences (PCC) across all scenes on these held-out points. Similar to PCK~\cite{yang2012articulated}, PCC considers a correspondence correct if the reprojected point lies within a radius defined as a percentage $alpha$ of the image size.
We run our method with various parameters and regularizations turned on and off for ablations and also compare against strong baselines like DUSt3R~\cite{wang2024dust3r} adapted to our setting, all shown in Fig.~\ref{fig:quantitative-baselines}. \new{In order for a fair comparison, we also compare with a version of DUST3R that uses our correspondences via adding $\mathcal{L}_{3D}$ in their global optimization stage along with our labeled masks applied to the confidence map. Results show that our proposed approach obtains the best PCC across all methods. We find that DUST3R works well on a few scenes, but when it fails, it fails catastrophically. Using our labeled correspondence help, but not significantly. Adding the dense alignment warp is necessary to significantly increase the performance. Our experiments validate the need for methods designed to deal with geometrically inconsistent input images.}

\vspace{-10px}

\begin{figure*}[t]
\centering
\includegraphics[width=\linewidth]{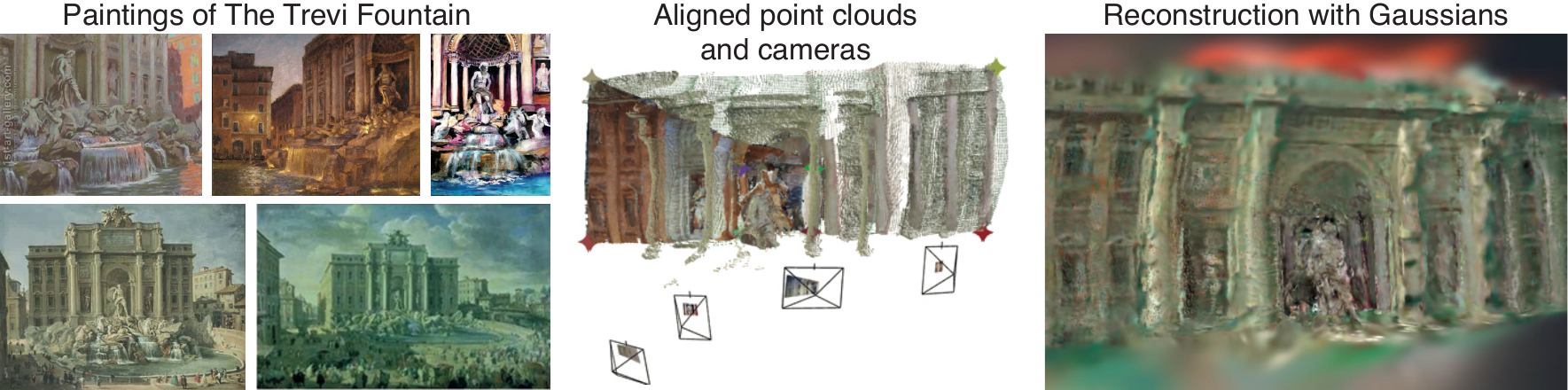}
\vspace{-15px}
\caption{\textbf{Reconstructing paintings with Toon3D.} Our method enables reconstructing paintings. On the left, we show a few paintings of The Trevi Fountain. In the middle, we show the recovered point cloud and cameras (with warped and cropped images). On the right, we densify the point cloud with Gaussian Splatting.}
\label{fig:paintings}
\vspace{-5mm}
\end{figure*}

\vspace{-5px}
\subsection{Sparse-view Reconstruction Validation}
\label{sec:sparse-view-reconstruction}
In this section, we validate the correctness of our approach on image collections that are geometrically consistent. Airbnb listings provide suitable test cases, as their photos are often geometrically consistent but sparse with wide baselines. For this evaluation, we reconstruct sparse photo collections from two Airbnb rooms from a listing (8 photos of a bedroom, shown in the project page, and 5 photos of a living room, shown in Fig.~\ref{fig:sparse-view-reconstruction}). This task is very difficult because SfM pipelines like COLMAP fail to find enough correspondences to accurately recover all poses. Furthermore, even with accurate camera poses, the sparse-view reconstruction setting is especially hard without priors or specialized methods like RegNeRF~\cite{niemeyer2022regnerf} or ReconFusion~\cite{wu2023reconfusion}. We tackle this sparse-view Airbnb setting with our method for two reasons: (1) to show that we can get COLMAP to work with labeled correspondences from the Toon3D Labeler and (2) to show that our approach works for real sparse photo collections, indicating applications beyond cartoons.

When running COLMAP on our Airbnb collections, default COLMAP only registers 46\% of the images. This could be possibly improved with better correspondences, \eg ~\cite{detone2018superpoint,tyszkiewicz2020disk,dusmanu2019d2}, but there is no guarantee of finding enough inlier correspondences if automated methods are used. With our Toon3D Labeler, however, we can manually label the images quickly and get COLMAP to succeed for all images. We compare the recovered COLMAP cameras with our correspondences with the cameras recovered from Toon3D. The mean relative rotation distance between corresponding pairs in our reconstructions vs. COLMAP's is quite low at only 8.29$^{\circ}$, indicating our cameras are similar to ones recovered by COLMAP with human-labeled correspondences. We do not compare translations or focal lengths due to ambiguity between the two, but we note that our camera relative rotations match COLMAP quite well, suggesting that our camera pose estimation is accurate.
We show qualitative results for sparse-view reconstruction on real images in Fig.~\ref{fig:sparse-view-reconstruction} and videos on the project page.

\subsection{Reconstructing paintings}
We also show our pipeline can reconstruct paintings of the same scene. Fig.~\ref{fig:paintings} shows results of Toon3D on paintings of The Trevi Fountain found in the Oxford Dataset~\cite{crowley2015search}. This setup requires multiple paintings of the same scene from diverse viewpoints, which is uncommon. However, it presents an interesting problem, and notably, we are able to apply the Toon3D pipeline successfully without any modifications.

\subsection{Visualizing inconsistencies}

\begin{figure}[H]
\centering
\includegraphics[width=\linewidth]{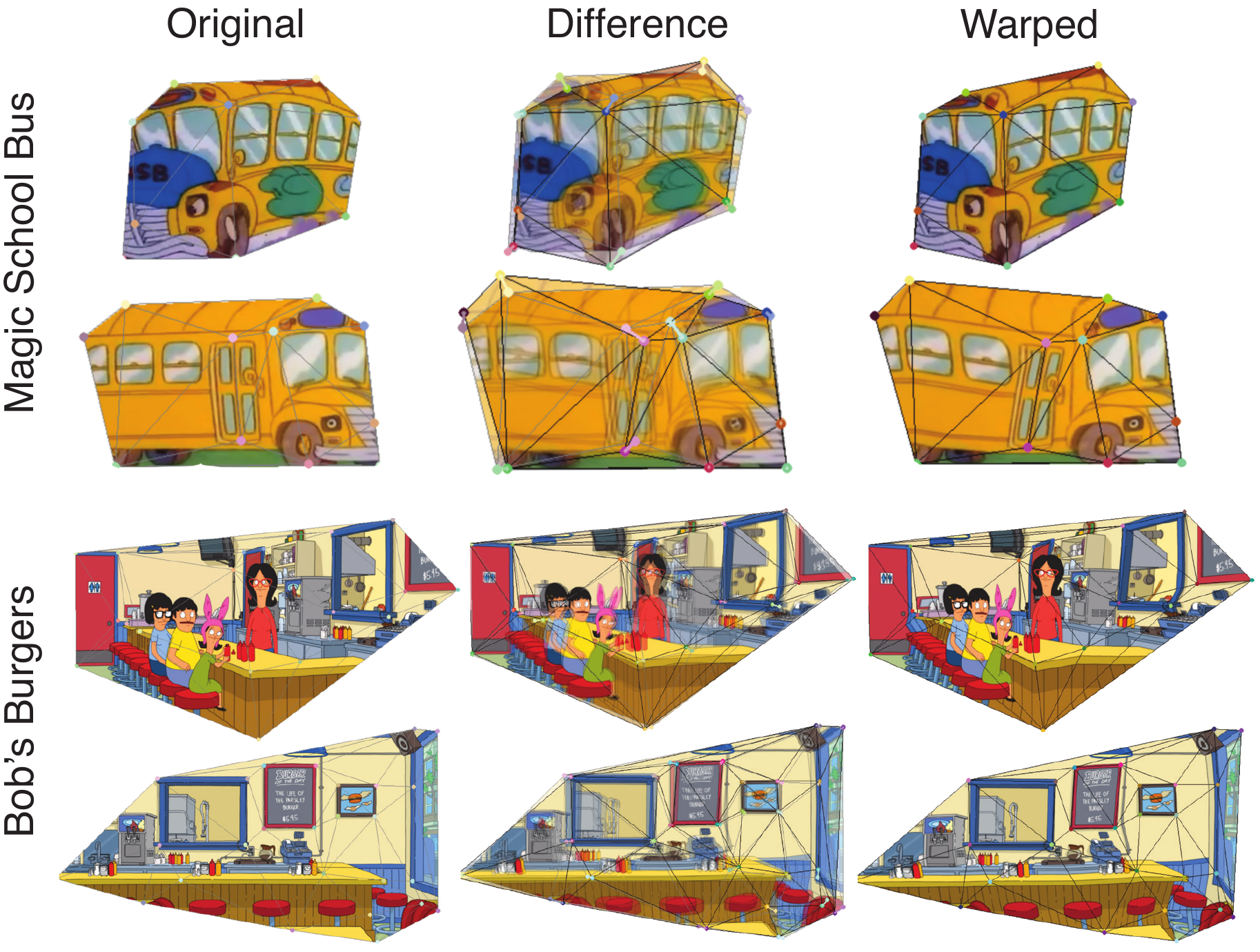}
\caption{\textbf{Visualizing inconsistencies.} We show the most inconsistent regions in a few images from different scenes by overlaying the original image (left) on top of the deformed image (right) to construct a difference image (middle). More blurry regions show where the images warped more to achieve 3D consistency.}
\label{fig:visualizing-inconsistencies}
\vspace{-1.5em}
\end{figure}

One unique aspect of Toon3D is that we keep the original images around rather than discarding them. They are warped in 2D to obey the global 3D consistency through a perspective camera model. This is fundamentally different than alternative sparse-view generative methods, \eg Dreambooth3D~\cite{raj2023dreambooth3d} which fine-tunes on a collection of images and then hallucinates a scene. In Fig.~\ref{fig:visualizing-inconsistencies} we show where the images deform the most to create a unified consistent 3D structure.
Additionally, it provides insights into the artistic techniques used to convey 3D or to emphasize regions in drawings without strictly adhering to physical laws.

\vspace{-1em}

\section{Conclusion}

We present Toon3D, a pipeline for 3D reconstruction from geometrically inconsistent images of a scene found in settings such as cartoons and animations. This is an interesting setup as humans have no problem interpreting the depicted scene in 3D, while as we show existing 3D reconstruction methods struggle in various ways. We propose a method that takes advantage of labeled correspondences and predicted depth priors to reconstruct these scenes by explaining away their inconsistencies by deforming the images to obey perspective projection models with regularizations. While our approach shows promising results, many exciting future directions remain, such as incorporating diffusion priors or data-driven methods to reconstruct cartoons end-to-end.
Finally, we encourage our method to be used ethically and responsibly when creating content for visual media.

\section*{Acknowledgements}

This project is supported in part by IARPA DOI/IBC 140D0423C0035. The views and conclusions contained herein are those of the authors and do not represent the official policies or endorsements of IARPA, DOI/IBC, of the U.S. Government. We would like to thank Qianqian Wang, Justin Kerr, Brent Yi, David McAllister, Matthew Tancik, Evonne Ng, Anjali Thakrar, Christian Foley, Abhishek Kar, Georgios Pavlakos, the Nerfstudio team, and the KAIR lab for discussions, feedback, and technical support. We also thank Ian Mitchell and Roland Jose for helping to label points.

%% file: evaluation_table.tex
\begin{table}[H]
\centering
\footnotesize
\setlength{\tabcolsep}{2.1pt}
\begin{tabular}{lc|lc}
\toprule
Method & PCC$\uparrow$ & Method & PCC$\uparrow$ \\
\midrule
\textit{Camera Alignment} &  & \textit{Deformation Alignment} & \\
$\mathcal{J}_{\text{camera}}$ & 0.26 & $\mathcal{J}_{\text{camera}} + \mathcal{J}_{\text{deform}}$ & 0.47  \\
$- \mathcal{L}_{focal}$ & 0.26 & $- \mathcal{L}_{z}$ & 0.42  \\
$- \mathcal{L}_{aspect}$ & 0.24 & $- \mathcal{L}_{ARAP_{2D}}$ & 0.42  \\
$- \mathcal{L}_{scale}$ & 0.18 & $-\mathcal{J}_{\text{deform}}$* & 0.36  \\
Traditional BA & 0.10 & Switch $\mathcal{L}_{3D}$ to $\mathcal{L}_{2D}$ & 0.31 \\
\bottomrule
\end{tabular}
\caption{\textbf{Quantitative ablations.} 
We report reprojection error for 5 holdout points on our 12 scenes. PCC is evaluated using a threshold radius set to 3\% of the image size ($\alpha=0.03$). 
(* Includes $\mathcal{L}_{flip}$)}
\vspace{-2.5em}
\label{tab:ablations}
\end{table}

%% file: appendix.tex
\section*{Appendix}

\begin{figure}[H]
\centering
\includegraphics[width=\linewidth]{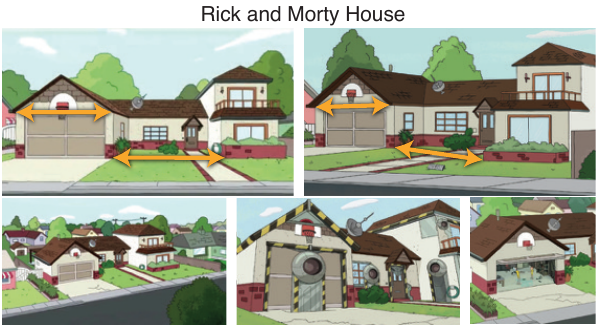}
\caption{\textbf{Geometrical inconsistencies in cartoons.} Are these orange arrows consistent? It is incredibly difficult to tell as a human, but COLMAP and SfM pipelines fail on these images, even with our hand-labeled correspondences.
}
\label{fig:motivation}
\end{figure}

Our Toon3D framework takes hand-drawn images with geometric inconsistencies and aligns them in 3D to create a consistent structure. Figure~\ref{fig:motivation} shows an example of a scene we work with.

\section{Video}

Our narrated video is available on our project webpage. It contains an overview of the paper and video results. It is complementary to our submitted PDF, which is composed of screen-captured and rendered frames. Our video results are more immersive than what 2D figures can convey.

\section{Baselines}

\begin{figure*}[t]
\centering
\includegraphics[width=\linewidth]{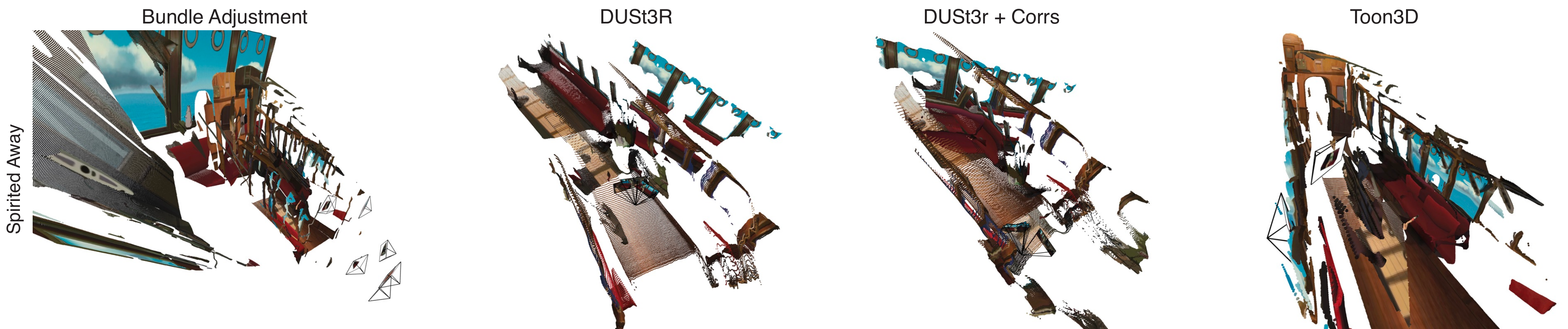}
\caption{\textbf{Baselines.} We compare our method on the Spirited Away scene with baselines mentioned in the paper.
Bundle Adjustment fails because it is unconstrained and doesn't use a prior to recover depth. We visualize the the result by backprojecting monocular depths at the recovered camera locations.
DUSt3R, a data-driven method, performs better and recovers a more plausible result but is still inconsistent.
DUSt3R + Corrs is sightly improved by using our labeled points at the correspondence locations, but it cannot recover fully from DUSt3R's initial prediction.
Toon3D (our method) produces the most consistent and realistic structure.
}
\label{fig:baselines}
\end{figure*}

We show a qualitative example of our baselines in \cref{fig:baselines}. Specifically, we show Bundle Adjustment, DUSt3R, DUSt3R + Corrs, and Toon3D (our method). DUSt3R + Corrs improves DUSt3R using our correspondences and 3D loss but it cannot reach the quality that we achieve with Toon3D. The PCC (as reported and explained in the paper) for each method on just the Spirited Away scene with 5 holdout correspondences is as follows for $\alpha = 0.05$: Bundle Adjustment (0.4), DUSt3R (0.1), DUSt3R + Corrs (0.2), and Toon3D (0.9) — higher is better. We achieve the best results qualitatively and quantitatively.

\section{Toon3D Labeler}

\begin{figure*}[t]
\centering
\includegraphics[width=\linewidth]{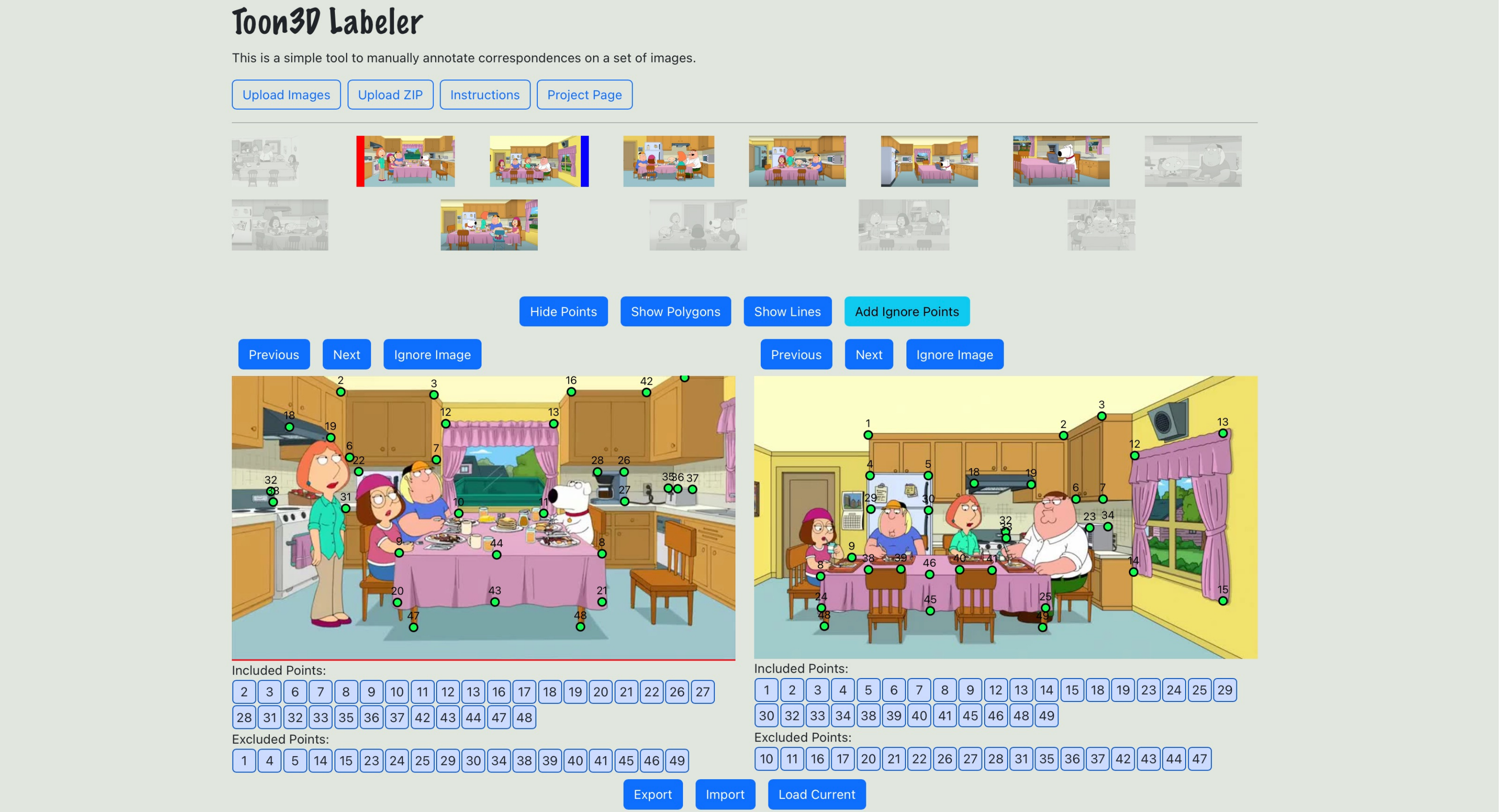}
\caption{\textbf{Toon3D Labeler.} Here is a screen capture from the Toon3D Labeler interface. Using the labeler, a user can label points and masks, and one can interactively visualize the depth map to avoid labeling on depth boundaries (see the overview video for a screen recording of this). Our Toon3D Labeler is a general labeling tool for labeling multi-view correspondences.}
\label{fig:toon3d-labeler}
\end{figure*}

Figure~\ref{fig:toon3d-labeler} shows a screen capture of the Toon3D Labeler. We will make this tool available for others to use.

\begin{figure*}[t]
\centering
\includegraphics[width=\linewidth]{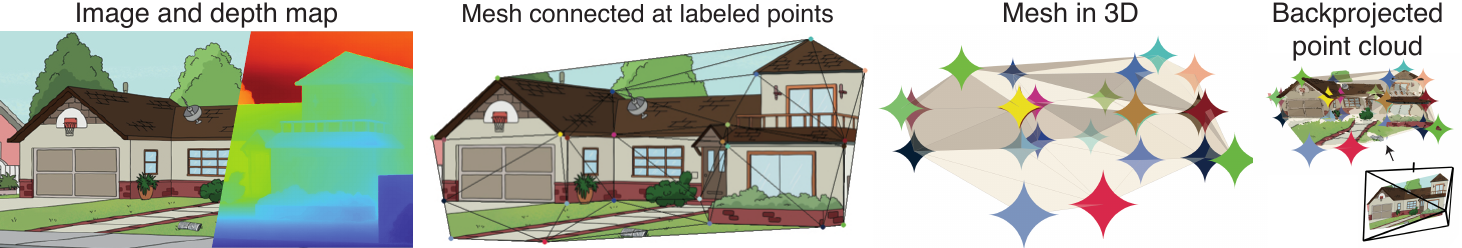}
\caption{\textbf{Deformable mesh topology.} We start with an image and predicted depth map (left). Then, we create a mesh with the 2D correspondences to define the topology (middle left). This mesh lives in 3D, where larger diamonds are closer to the camera (middle right). We optimize the 3D vertices to achieve multi-view consistency. After convergence, we use barycentric interpolation to query the RGB and depth maps in order to create the dense 3D point cloud, shown on the right.}
\label{fig:topology}
\end{figure*}

\section{Toon3D Dataset}

We choose to use cartoon scenes that are hand-drawn rather than using animated scenes that are rendered or based on an underlying 3D model. We select a variety of cartoons based on popularity. Table~\ref{tab:dataset} shows our datasets and relevant annotation info, including how many images we use to create each scene and how many point labels are used. We use a varying number of point labels, ranging from only 46 points (Magic School Bus) to as many as 191 points (BoJack Room) in a particular scene. This range is meant to convey the robustness of our method to handle a few or many user-defined correspondences. Our Toon3D Labeler will be released so others can label scenes as they desire.

\setlength{\tabcolsep}{5pt} 
\begin{table}[H]
\caption{\textbf{Toon3D Dataset.} Here are some statistics for the Toon3D Dataset. We have $\sim$7 images per scene, for a total of 79 images across the 12 scenes. Each image has on average 18.3 points per image, but it varies per scene.}
  \label{tab:dataset}
    \centering
    \captionsetup{font=scriptsize}
    \scriptsize
    \begin{tabular}{l|cc|c} 
        \toprule
 & Num images & Num points & Avg. num points / image  \\
        \midrule
Avatar House & 8 & 156 & 19.5 \\
Bob's Burgers & 7 & 147 & 21.0 \\
BoJack Room & 12 & 191 & 15.9 \\
Family Guy Dining & 7 & 184 & 26.3 \\
Family Guy House & 6 & 133 & 22.2 \\
Krusty Krab & 9 & 82 & 9.11 \\
Magic School Bus & 5 & 46 & 9.20 \\
Mystery Machine & 6 & 55 & 9.17 \\
Planet Express & 5 & 137 & 27.4 \\
Simpsons House & 5 & 137 & 27.4 \\
Rick and Morty & 4 & 99 & 24.8 \\
Spirited Away & 5 & 75 & 15.0 \\ \hline\noalign{\smallskip}
 Total & 79 & 1442 & 18.3 \\
        \bottomrule
    \end{tabular}
    \label{tab:counts}
\end{table}

\section{Deformable mesh topology}

In \cref{fig:topology}, we show an illustration of how we go from an image, a depth map, and our labeled correspondences, to a 3D mesh which can be deformed.

\section{Sparse-view reconstruction data}

We obtain sparse-view images from Airbnb from this listing: \url{https://www.airbnb.com/rooms/833261990707199349}. Our overview video shows the two rooms and their images. The ``Living room'', shown in the paper as well, has 5 images. ``Bedroom 2" has 8 images. Videos of our Toon3D reconstructions and renders are shown for both rooms in our overview video.